\documentclass[conference]{IEEEtran}
\IEEEoverridecommandlockouts
\usepackage{cite}
\usepackage{amsmath,amssymb,amsfonts}
\usepackage{algorithmic}
\usepackage{graphicx}
\usepackage{textcomp}
\usepackage{xcolor}
\usepackage{amsthm, algorithm, float, geometry, breakcites, enumitem, url, booktabs, arydshln}
\usepackage{mathrsfs}

\def\BibTeX{{\rm B\kern-.05em{\sc i\kern-.025em b}\kern-.08em
    T\kern-.1667em\lower.7ex\hbox{E}\kern-.125emX}}
\begin{document}

\title{PINNs-Based Uncertainty Quantification for Transient Stability Analysis\\
}


\author{\IEEEauthorblockN{Ren Wang\textsuperscript{a}, Ming Zhong\textsuperscript{b}, Kaidi Xu\textsuperscript{c}, Lola Giráldez Sánchez-Cortés\textsuperscript{a}, Ignacio de Cominges Guerra\textsuperscript{a}}
\IEEEauthorblockA{\textsuperscript{a}\textit{Department of Electrical and Computer Engineering, Illinois Institute of Technology, Chicago, IL 60616 US} \\
\textsuperscript{b}\textit{Applied Mathematics Department, Illinois Institute of Technology, Chicago, IL 60616 US} \\
\textsuperscript{c}\textit{Department of Computer Science, Drexel University, Philadelphia, PA 19104 US}\\
}
\thanks{This work is supported by the US National Science Foundation under Grant $2319243$, the ORAU Ralph E. Powe Junior Faculty Enhancement Award, and AoF-$2225507$. Corresponding: Ren Wang (\text{rwang74}@iit.edu)}
}

\maketitle

\begin{abstract}
This paper addresses the challenge of transient stability in power systems with missing parameters and uncertainty propagation in swing equations. We introduce a novel application of Physics-Informed Neural Networks (PINNs), specifically an Ensemble of PINNs (E-PINNs), to estimate critical parameters like rotor angle and inertia coefficient with enhanced accuracy and reduced computational load. E-PINNs capitalize on the underlying physical principles of swing equations to provide a robust solution. Our approach not only facilitates efficient parameter estimation but also quantifies uncertainties, delivering probabilistic insights into the system behavior. The efficacy of E-PINNs is demonstrated through the analysis of $1$-bus and $2$-bus systems, highlighting the model's ability to handle parameter variability and data scarcity. The study advances the application of machine learning in power system stability, paving the way for reliable and computationally efficient transient stability analysis.
\end{abstract}


\section{Introduction}
Machine learning has significantly influenced power system stability and dynamics for quite some time \cite{alimi2020review}. Recent advancements include the ever-increasing load demands, integration of renewable energy, and distributed energy resources (DERs), substantially altering the power systems' landscape and raising stability and security concerns \cite{wu2023transient,wang2014small}. This shift necessitates accurate transient simulations and marks the obsolescence of viewing networks as static entities, driving the demand for precise simulation tools for comprehensive stability studies and ensuring system reliability \cite{alsharief2019transient}.

Swing equations, describing synchronous generator dynamics, play a critical role in power system transient stability. They posed a significant challenge due to the computation intensity nature for solving the accompanying differential equations through traditional techniques such as Euler’s, Adams’, Runge-Kutta, and Newton-Raphson \cite{glover2012power}. The transition to variable renewable energy sources complicates this further, introducing swift and severe transients. To mitigate these issues, data-driven approaches have been explored, offering rapid inference speeds and excellent performance with ample data \cite{alimi2020review, sun2019independence, zhang2018review}. However, they require large data sets, are susceptible to data contamination, and may struggle with predicting extreme scenarios. A potential solution is to forge a direct link to the foundational physical principles. Fortunately, a fusion of prior physics-based knowledge with machine learning algorithms has resulted in a new machine learning method, namely the Physics-Informed Neural Networks (PINNs), marking a leap forward in accuracy and reliability of the models in use \cite{cuomo2022scientific}. In situations where data is scarce, computational speed is paramount, and there is a robust understanding of the theoretical model, PINNs present a promising mean for addressing differential equations in power system transient stability. PINNs have found applications across various domains of power systems, including solving AC power flows \cite{powerflows} and analyzing converter dynamics \cite{misyris2021capturing}, as well as in transient stability analysis \cite{misyris2020physics, stiasny2021transient}. 

However, challenges remain, particularly in relation to unknown parameters. Parameters in swing equations can be uncertain or unknown in numerous scenarios, such as in customized or modified equipment that might have parameters differing from standard models. Additionally, some parameters may vary based on operational conditions, and these dependencies might not be fully understood or accurately modeled. While additional data could potentially be utilized to determine the correct parameters, the presence of noise from various sources, including environmental conditions and device limitations, complicates the procure of accurate data extraction \cite{wang2018data,gao2018low}. The noisy or corrupted data collected for PINNs could lead to unreliable predictions.

Our research is poised to bridge these gaps, with a focus on estimating the rotor angle and inertia coefficient. In this study, we exploit the capabilities of neural networks and integrate them with the physical constraints inherent in the swing equation to efficiently estimate these parameters. Our research stands out in its ability to showcase the effectiveness of PINNs in addressing power system transient stability challenges and in quantifying the uncertainty surrounding the inertia coefficient. Accurately determining the inertia coefficient is challenging due to a multitude of factors including limited data availability, variations among generators, time-dependent changes, and intrinsic modeling uncertainties. Our approach involves employing an ensemble of PINNs (E-PINNs), allowing us to encompass uncertainty and provide probabilistic assessments of power system behavior and performance. Consequently, we can deliver both the variance of the Gaussian distribution of the predicted parameter and the confidence interval, highlighting the accuracy of our estimated inertia coefficient. The practical applicability of our method is demonstrated through examples involving 1-bus and 2-bus systems.
\section{Problem Formulation: Swing Equations With Unknown Parameters}\label{sec: PST}

The stability of a power system is a pivotal concern that encapsulates the system's capability to remain in a state of equilibrium, and to endure not only during normal operations but also when subjected to abrupt disruptions. The component of stability known as transient stability is particularly concerned with the system's ability to return to a steady state after sudden occurrences like changes in load, faults, or generator outages. For a more granular understanding of transient stability, professionals frequently resort to employing the swing equation. This fundamental equation is instrumental in assessing the dynamic response of three-phase synchronous generators during transient disturbances. It effectively encapsulates the movement of rotor angles, providing a streamlined yet profound view of the system's initial reaction to stability perturbations. Despite its inherent approximations, the swing equation is a significant analytical resource for initial evaluations of a power system's response to perturbations. The behavior of the rotor angles $\delta_i$ for generator $i$ is modeled by the swing equation as shown in \eqref{eq:SE 1.2}:

\begin{equation}
\label{eq:SE 1.2}
m_{i} \dfrac{d^2\delta_i}{dt^2} + d_i \frac{d\delta_i}{dt} + P_{ei} = P_{mi},
\end{equation}
Here, $m_i$ symbolizes the inertia coefficient for generator $i$, $d_i$ represents the damping coefficient, $P_{ei}$ is indicative of the electrical power output which includes electrical losses, while $P_{mi}$ corresponds to the mechanical power input from the prime mover, accounting for mechanical losses, all expressed in per unit values. To determine the electrical power at each bus $P_{ei}$ from \eqref{eq:SE 1.2}, it is necessary to address the power flow problem given by \eqref{eq:PF 1}:

\begin{equation}
\label{eq:PF 1}
P_{ei} = V_{i} \sum_{n=1}^{N} b_{in} V_n  \cos(\theta_i - \theta_n - \theta_{Y_{in}}) + P_{li}
\end{equation}
In this expression, $b_{in}$ stands for the susceptance values within the system matrix. The angles $\theta_{i}$, $\theta_{n}$, and $\theta_{Y_{in}}$ pertain to the angular positions at buses $i$ and $n$, and the angle of the line impedance, respectively. $V_{i}$ and $P_{li}$ denote the voltage magnitude and the load at each bus.

In various situations, the parameters within the swing equations can be rife with uncertainty or remain unidentified, such as in bespoke or modified equipment where the parameters deviate from standard specifications. Moreover, certain parameters might fluctuate with operational conditions, and these changes are often not completely comprehended or precisely replicated in models. While supplementary data could be harnessed to pinpoint accurate parameters, the precision of measurement devices and sensors is not absolute, resulting in parameter uncertainties. The data afflicted by noise or corruption which is gathered for inverse problem methods, including PINNs, can lead to untrustworthy forecasts. Within this complex landscape, we embark on a dual-pronged inquiry aimed at overcoming two main obstacles. Our first endeavor is to predict the outcome of the swing equation: the rotor angle $\delta_i$ in \eqref{eq:SE 1.2}. The second challenge lies in deducing the inertia coefficient $m_i$ for our generators, a parameter that remains elusive within our swing equation. The ambiguous measurements and the inherent data acquisition uncertainties add layers of complexity to our grasp of the system's operations. These dual difficulties — the elusive parameters and the pervasive uncertainty — overshadow the analysis of power system stability, compelling the exploration of advanced methods beyond conventional approaches.

\section{Our Method}\label{sec: pinns}
\subsection{PINNs on Power Systems}
In this section we delve into the architecture and methodology of Physics-Informed Neural Networks (PINNs) on power systems. A PINN on power systems comprises four main components: the input, the neural network, the output, and the loss function. The input domain of swing equations, denoted as $\Omega = [0, T] \times [0, P]$, corresponds to time $t$ and the input mechanical power $P_{mi}$ ranges specified in the dataset subsection. The neural network input itself is bifurcated into two distinct terms: $(t_u, P_u, \bar{u})$ is data-driven, while $(t_f, P_f)$ is derived from a theoretical physical model, which is defined by ODEs/PDEs. Initially, $t_u$ and $P_u$ in the data-driven part are time and power data samples from initial/boundary conditions, and $\bar{u}$ corresponds to the ground truth of $(t_u, P_u)$. However, further in our study we will add a certain percentage of labeled collocation points in the supervised learning part, which consists of data placed within the domain of our problem. The total number of labeled data is denoted by $N_u$. The variables $t_f$ and $P_f$ represent the time and power in the general region for a set of points $N_f$ within the domain. These points are subjected to validation by the theoretical physical model, specifically the swing equation of the particular system. The set of differential equations for each bus will be denoted as $f(t,P)$, following the formulation presented in \eqref{eq:DE 1}, aligning with our observations from \eqref{eq:SE 1.2}. 

\begin{equation}
\label{eq:DE 1}
 f(t,P) = \dfrac{2H_{i}}{w_0}\dfrac{d^2\delta_i}{dt^2} + P_{ei}  + D_i \frac{d\delta_i}{dt} - P_{mi}
\end{equation}



Overall, the ultimate objective of the neural network is to discover the most suitable $\theta$ parameters while aiming to minimize $\mathcal{L}_{MSE}$: 
\begin{equation}
\label{LF 1}
\begin{aligned}
    &\mathcal{L}_{M S E} =\mathcal{L}_{M S E, u}+\mathcal{L}_{M S E, f}, \quad \text{where} \\&\mathcal{L}_{M S E, u} = \frac{1}{N_u} \sum_{k=1}^{N_u}\left|u\left(t_u^k, x_u^k\right)-\bar{u}^k\right|^2\\& \mathcal{L}_{M S E, f} = \frac{1}{N_f} \sum_{j=1}^{N_f}\left|f\left(t_f^j, x_f^j\right)\right|^2
\end{aligned}
\end{equation}
where $\mathcal{L}_{MSE, u}$ corresponds to the loss of labeled data and $\mathcal{L}_{MSE, f}$ corresponds to the differential equation loss.

\subsection{Uncertainty Quantification} \label{UQ}
The operation of quantifying and assessing uncertainties that develop in various power system components, such as generation, transmission, and distribution, is known as uncertainty quantification (UQ). UQ methods in power systems typically involve statistical analysis, probabilistic modeling, and simulation-based approaches. These techniques help assess the likelihood and potential impacts of different uncertainties on power system variables such as voltage levels, power flows, stability, and reliability. In this section we will consider the inertia, $m_{i}$ as the unknown parameter of our equation. We will then quantify its uncertainty providing the variance and average of its Gaussian distribution given a set of time and power. Finally we will calculate the confidence interval, which will help us estimate how accurate or estimation is.

Inspired by \cite{jiang2022pinn}, we employ the Ensemble PINNs (E-PINN) to achieve the above objectives. The framework for each bus contains $n$ pairs of fully connected neural networks, where each pair of neural networks has one neural network predicting the rotor angle and a second neural network predicting its inertia coefficient. Each PINN member of the E-PINN relies on model diversity to generate uncertainty estimates. There are $n$ number of models, each of them shows variation from the other as they are trained with a different range noise in the data, different boundary conditions for power and time, different percentages of collocation points for the supervised learning part, and each of them uses different initialization for all network weights and biases.

The last step involves the inversion of the Quantity of Interest (QoI), which is achieved through the posterior Gaussian distribution $p(m_{i}|t,P,u)$. Once the E-PINN has been trained, a surrogate model for the QoI has been created. Only the inputs are relevant to the QoI prediction. Our estimate inertia coefficient can be approximated by $\hat{m}_i(t,P;\theta_j)$ for the neural network $j$:

\begin{equation}
\begin{aligned}
& p(m_{i}|x,d) \sim \mathcal{N}(m_{i}(t,P), \sigma^2) \\
& \mu \approx n^{-1} \sum_{j=1}^{n} \hat{m}(t,P; \theta_{j}) \\
& \sigma^2 \approx n^{-1} \sum_{j=1}^{n} (\hat{m}(t,P; \theta_{j})^2-\mu^2)
\end{aligned}
\end{equation}



The equation quantifies the QoI uncertainty by applying ensemble statistics, as it procures the mean and variance of its Gaussian Distribution, where $\theta_{j}$ represents the parameters of the $j$-th model. By employing ensemble statistics, the equation provides an unbiased estimate of the mean of the QoI. This approach naturally incorporates the uncertainty inherent to the model itself.

\section{Experimental Results}\label{ESR}
The MATLAB numerical solver \textit{ode$45$} was employed to generate the dataset, creating $100$ varied trajectories for power variation and documenting $201$ time steps for each trajectory. As a result, the final datasets encompassed $20100$ angle data samples. Simulations were conducted for a $1$-bus system with a single generator, and a $2$-bus system with two generators and one load. For training our E-PINN, six different models for the $1$-bus and three for the $2$-bus system were utilized.

\begin{figure}[ht]
		\centering
	\includegraphics[width=1.0\linewidth]{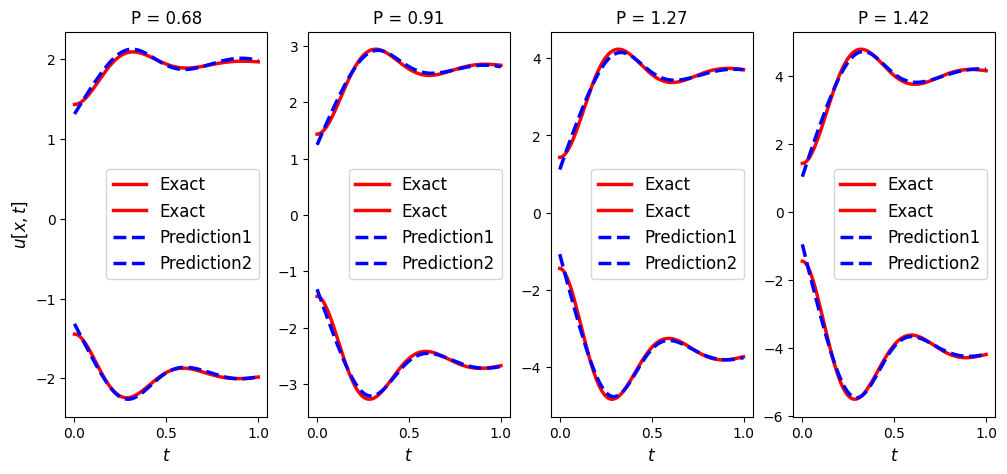}
		\caption{Comparison of the exact solution with the PINN predicted angle $\delta$ in a 2-bus system.
		}
		\label{2Gen}
\end{figure}
Uncertain active power input was assumed within the ranges $[0.08, 0.18]$ for the $1$-bus system and $[0.51, 1.51]$ for the load in bus $2$ in the $2$-bus system. for the load in the $2$-bus system. A hundred trajectories over time intervals $[0, 20]s$ and $[0, 1]s$ were generated for the $1$-bus and $2$-bus systems respectively. Voltage magnitudes $V_{1}$ and $V_{2}$ in the $2$-bus system were maintained at $1$ p.u. with $b_{12}= 0.2$ p.u. The inertia coefficient was treated as an unknown in the ODE loss. For the unsupervised learning phase, $9000$ out of the possible $20100$ points were used. Diverse noise levels were introduced to the input data across different models, and additional labeled data ranged from $25\%$ to $50\%$. We also adjusted the range's upper and lower bounds for power change in each model. Figure~\ref{2Gen} illustrates the angle predictions of the two generators in the 2-bus system by the PINN, displaying the predicted angles in blue and the actual angles in red, across four distinct power trajectories. Notably, the angle estimations are precise, with average absolute relative errors of approximately $3.420585 \times 10^{-3}$ and $3.486263 \times 10^{-3}$ for generators $1$ and $2$, respectively.

\begin{figure}[ht]
\begin{minipage}[b]{.49\linewidth}
  \centering
  \centerline{\includegraphics[trim=0 0 0 0,clip,width=1.0\textwidth]{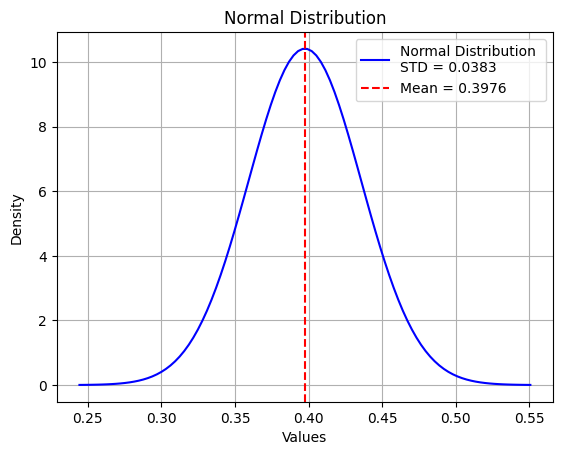}}
  \centerline{(a)}\medskip
\end{minipage}
\begin{minipage}[b]{.49\linewidth}
  \centering
  \centerline{\includegraphics[trim=0 0 0 0,clip,width=1.0\textwidth]{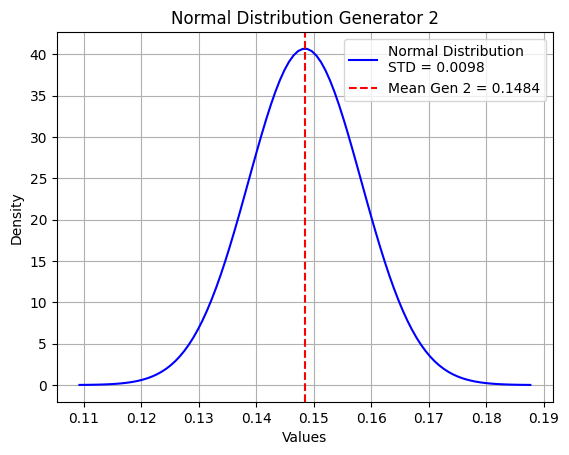}}
  \centerline{(b)}\medskip
\end{minipage}
\caption{(a) Gaussian Distribution of the inertia parameter obtained by the $1$-bus system E-PINN with Mean $0.3976$ and Standard Deviation $0.0383$; (b) Gaussian Distribution of the inertia parameter of Generator $2$ obtained by the $2$-bus system E-PINN with Mean $0.1484$ and Standard Deviation $0.0098$.}
\label{GaussGen}
\end{figure}

Subsequent to deploying Ensemble PINNs, the resulting Gaussian Distributions for generators $1$ and $2$ are depicted in Figure~\ref{GaussGen}. Analysis of these distributions reveals mean inertia coefficients of $0.3976$ and $0.1484$ for the swing equation, closely matching the ground truth values of $0.4$ for the $1$-bus system and $0.132629$ for generator $2$ in the $2$-bus system, respectively. 

\section{Conclusion}\label{Concl} 

In conclusion, our study has successfully demonstrated the efficacy of Physics-Informed Neural Networks (PINNs) in the realm of power system stability analysis, particularly in transient stability assessment with uncertain and missing parameters. By leveraging the ensemble of PINNs (E-PINNs), we have presented a method that not only predicts key parameters such as rotor angle and inertia coefficient with remarkable precision but also quantifies the associated uncertainty through probabilistic measures. Future work can explore the adaptation of this methodology to uncertainty of rotor angels, damping coefficients, a broader range of power system models, and the inclusion of real-world operational data, with the ultimate goal of achieving a more resilient and reliable power grid.

\bibliographystyle{ieeetr}
\bibliography{refs/ref_mr,refs/ref_new,refs/ref_stability}

\vspace{12pt}

\end{document}